\documentclass[graybox]{svmult}

\usepackage{mathptmx}       
\usepackage{helvet}         
\usepackage{courier}        
\usepackage{type1cm}        
\usepackage{makeidx}         
\usepackage{graphicx}        
\usepackage{multicol,url}        
\usepackage[bottom]{footmisc}

\makeindex             

\begin{document}


\titlerunning{Information-theoretic and Algorithmic Approach to Cognition}
\title*{The Information-theoretic and Algorithmic Approach to Human, Animal and Artificial Cognition}
\author{Nicolas Gauvrit, Hector Zenil\thanks{First two authors contributed equally.} and Jesper Tegn\'er}
\institute{Hector Zenil \at Unit of Computational Medicine, Department of Medicine Solna\\
Centre for Molecular Medicine \& SciLifeLab, Karolinska Institutet, Stockholm, Sweden \& \\
Department of Computer Science, University of Oxford, UK\\Corresponding author: \email{hector.zenil@algorithmicnaturelab.org}
\and Nicolas Gauvrit \at Human and Artificial Cognition Lab\\Universit\'e Paris 8 \& EPHE, Paris, France.\\
\email{ngauvrit@me.com}
\and Jesper Tegn\'er \at Unit of Computational Medicine, Department of Medicine Solna\\
Centre for Molecular Medicine \& SciLifeLab, Karolinska Institutet, Stockholm, Sweden.\\ \email{jesper/tegner@ki.se}}
%
%
\maketitle

\abstract*{We survey concepts at the frontier of research connecting artificial, animal and human cognition to computation and information processing---from the Turing test to Searle's Chinese Room argument, from Integrated Information Theory to computational and algorithmic complexity. We start by arguing that passing the Turing test is a trivial computational problem and that its pragmatic difficulty sheds light on the limits of the computational nature of the human mind and thereby its necessary algorithmic nature more than it does on the challenge of artificial intelligence. We then review our proposed algorithmic information-theoretic measures for quantifying and characterizing cognition in various forms. These are capable of accounting for known biases in human behavior, thus vindicating a computational algorithmic view of cognition as first suggested by Turing, but this time rooted in the concept of algorithmic probability, which in turn is based on computational universality while being independent of computational model, and which has the virtue of being predictive and testable as a model theory of cognitive behavior.}

\abstract{We survey concepts at the frontier of research connecting artificial, animal and human cognition to computation and information processing---from the Turing test to Searle's Chinese Room argument, from Integrated Information Theory to computational and algorithmic complexity. We start by arguing that passing the Turing test is a trivial computational problem and that its pragmatic difficulty sheds light on the computational nature of the human mind more than it does on the challenge of artificial intelligence. We then review our proposed algorithmic information-theoretic measures for quantifying and characterizing cognition in various forms. These are capable of accounting for known biases in human behavior, thus vindicating a computational algorithmic view of cognition as first suggested by Turing, but this time rooted in the concept of algorithmic probability, which in turn is based on computational universality while being independent of computational model, and which has the virtue of being predictive and testable as a model theory of cognitive behavior.}

\section{The algorithmic model of mind}

Judea Pearl, a leading theorist of causality, believes that every computer scientist is in a sense a frustrated psychologist, because computer scientists learn about themselves (and about others) by writing computer programs that are emulators of intelligent human behavior~\cite{pearl}. As Pearl suggests, computer programs are in effect an enhanced version of an important intellectual component of ourselves: they perform calculations for us, albeit not always as we ourselves would perform them. Pearl claims that by emulating ourselves we externalize a part of our behavior; we mirror our inner selves, allowing them to become objects of investigation. We are able to monitor the consequences of changing our minds by changing the code of a computer program, which is effectively a reflection of our minds.

Perhaps Alan Turing was the first such frustrated psychologist, attempting to explain human behavior by means of mechanical processes, first in his seminal work on universal computation~\cite{turing}, but also in his later landmark paper connecting intelligence and computation through an imitation game~\cite{turing2}. On the one hand, Artificial Intelligence has sought to automatize aspects of behavior that would in the past have been considered intelligent, in the spirit of  Turing's first paper on universal computation. It has driven the evolution of computer programs from mere arithmetic calculators to machines capable of playing chess at levels beyond human capacity and answering questions at near---and in some domains beyond---human performance, to machines capable of linguistic translation and face recognition at human standards. On the other hand, the philosophical discussion epitomized by Turing's later paper~\cite{turing2} on human and machine intelligence prompted an early tendency to conflate intelligence and consciousness, generating interesting responses from scholars such as John Searle. Searle's Chinese Room argument~\cite{searle} (CRA) is not an objection against Turing's main argument, which has its own virtues despite its many limitations, but a call to distinguish human consciousness from intelligent behavior in general. Additionally, the philosophy of mind has transitioned from materialism to functionalism to computationalism~\cite{gordana}, but until very recently little had been done by way of formally---conceptually and technically---connecting computation to a model of consciousness and cognition.

Despite concerns about the so-called \textit{Integrated Information Theory} (IIT)---which by no means are devastating or final even if valid~\footnote{See e.g. \url{http://www.scottaaronson.com/blog/?p=1799} as accessed on December 23, 2015. Where Tononi himself provided acceptable, even if not definite, answers.}---there exists now a contending formal theory of consciousness, which may be right or wrong but has the virtue of being precise and well-defined, even though it has evolved in a short period of time. Integrated Information Theory~\cite{tononi} lays the groundwork for an interesting computational and information-theoretic account of the necessary conditions for consciousness. It proposes a measure of the integration of  information between the interacting elements that account for what is essential to consciousness, viz. the feeling of an internal experience, and therefore the generation of information within the system in excess of information received from the external environment and independent of what the system retrieves, if anything. 

Moreover, there now exists a formal theory capable of accounting for biases in human perception that classical probability theory could only quantify but not explain or generate a working hypothesis about~\cite{gauvrit2014a, gauvrit2014c}. This algorithmic model provides a strong formal connection between cognition and computation by means of recursion, which seems to be innate or else developed over the course of our lifetimes~\cite{gauvrit2014c}. The theory is also not immune to arguments of super-Turing computation~\cite{zenilmind}, suggesting that while the actual mechanisms of cognition and the mind may be very different from the operation of a Turing machine, its computational power may be that of the Turing model. But we do not yet know with certainty what conditions would be necessary or sufficient to algorithmically account for the same mental biases with more or less computational power, and this is an interesting direction for further investigation. More precisely, we need to ascertain whether theorems such as the algorithmic Coding theorem are valid under conditions of super- or subuniversality. Here it is irrelevant, however, whether the brain may look like a Turing machine, which is a trivialization of the question of its computational power because nobody would think the brain operates like a Turing machine.

The connections suggested between computation and life go well beyond cognition. For example, Sydney Brenner, one of the fathers of molecular biology, argues that Turing machines and cells have much in common~\cite{brenner} even if these connections were made in a rather top level fashion reminiscence of old attempts to trivialize biology. More recently, it has been shown in systematic experiments with yeast that evolutionary paths taken by completely different mutations in equivalent environments reach the same evolutionary outcome~\cite{kryazhimskiy}, which may establish another computational property known as \textit{confluence} in abstract rewriting systems, also known as the \textit{Church-Rosser property} (of which the Turing machine model is but only one type of rewriting system). This particular kind of contingency makes evolution more predictable than expected, but it also means that finding paths leading to a disease, such as a neurodegenerative disease, can be more difficult  because of the great diversity of possible causes leading to the same undesirable outcome. Strong algorithmic connections between animal behavior, molecular biology and Turing computation have been explored in~\cite{entropy,zenilubiquity}. But the connections between computation and cognition can be traced back to the very beginning of the field, which will help us lay the groundwork for what we believe is an important contribution towards a better understanding of cognition, particularly human cognition, through algorithmic lenses. Here we pave the first steps towards revealing a stronger non-trivial connection between computation (or information processing) on one side and cognition in the other side by means of the theory of algorithmic information.

\subsection{The Turing test is trivial, ergo the mind is algorithmic}

In principle, passing the Turing test is trivially achievable by brute force. This can be demonstrated using a simple Chinese Room Argument-type thought experiment. Take the number of (comprehensible) sentences in a conversation. This number is finite because the set of understandable finite sentences is bounded and is therefore finite. Write a \textit{lookup table} with all possible answers to any possible question. A thorough introduction to these ideas is offered in~\cite{drew}. Passing the Turing test is then trivially achievable in finite time and space by brute force; it is just a combinatorial problem but nobody would suspect the brain operating in this way. Lookup tables run in $O(1)$ computational time, but if a machine is to pass the Turing test by brute-force, their size would grow exponentially for sentences that only grow linearly, given the wide range of possible answers. Passing the Turing test, even for conservative sentence and conversation lengths, would require more space than the observable universe.



One can make the case of certain sentences of some infinite nature that may be understood by the human mind. For example, we can build a nested sentence using the word \textit{``meaning"}. \textit{``The meaning of meaning"} is still a relatively easy sentence to grasp, the third level of nesting, however, may already be a challenge, and if it is not, then one can nest it $n$ times until the sentence appears beyond human understanding. At some point, one can think of a large $n$ for which one can still make the case that some human understanding is possible. One can make the case for an unbounded, if not infinite $n$, from which the human mind can still draw some meaning making it impossible for a lookup table implementation to deal with, because of the unbounded, ever increasing $n$. This goes along the lines of Hofstadter self-referential loops~\cite{hofstadter}, think of the sentence \textit{``What is the meaning of this sentence"} which one can again nest several times, making it a conundrum but from which an arbitrary number of ``nestings" the human mind may still be able to grasp something, even if a false understanding of its meaning, and maybe even collapsing on its own making some sort of \textit{strange loop} where the larger the $n$ the less absolute information content in the sentence there is, not only relative to its size, hence approaching 0 additional meaning for ever increasing large $n$, hence finite meaning out of an asymptotically infinite nested sentence. Understanding these sentences in the described way seems to require ``true consciousness'' of the nature of the context in which these sentences are constructed similar to arguments in favor of ``intuiton''~\cite{penrose} spanning different levels of understanding (inside and outside the theory) leading and based upon G{\"o}del's~\cite{godel} type arguments.

While it is true that not all combinations of words form valid grammatical sentences, passing the Turing test by brute force may also actually require the capacity to recognize invalid sentences in order either to avoid them or find suitable answers for each of them. This conservative number of combinations also assumes that for the same sentence the lookup table produces the same answer, because the same sentence will have assigned to it the same index produced by a hash function. This amounts to implementing a lookup table with no additional information--such as the likelihood that a given word will be next to another--thus effectively reducing the number of combinations. But assuming a raw lookup table with no other algorithmic implementation, in order to implement some sort of memory it would need to work at the level of conversations and not sentences or even paragraphs. Then the questioner can have a reasonable answer to the question \textit{``Do you remember what my first question was?''} as they must in order to be able to pass a well-designed Turing test. This means that in order to answer any possible question related to previous questions the lookup table has to be simply astronomically even larger, and not that a lookup table approach cannot pass the Turing test. While the final number of combinations of possible conversations, even using the most conservative numbers, is many orders of magnitude larger than the largest astronomical magnitudes, this is still a finite number. This means that, on the one hand, the Turing test is computationally trivially achievable because one only needs to build a large enough lookup table to have an answer for each possible sentence in a conversation. On the other hand, given both that the Turing test is in practice impossible to pass by brute force using a lookup table, and that passing the test is in fact achievable by the human mind, it cannot be the case that the mind implements a lookup table~\cite{kirk,perlis}. The respective ``additional mechanisms'' are the key to the cognitive abilities.

And this is indeed what Searle helped emphasize. Objections to the Turing test may be of a metaphysical stripe, or adhere to Searle and introduce resource constraints. But this means that either the mind has certain metaphysical properties that cannot be represented and reproduced by science, or that the Turing test and therefore the computational operation of the mind can only make sense if resources are taken into account~\cite{parberry,dowe1,dowe2}. Which is to say that the Turing test must be passed using a certain limited amount of space and in a certain limited time, and if not, then machine intelligence is unrelated to the Turing test (and to computing), and must be understood as a form of rule/data compression and decompression. Searle is right in that the brain is unlikely to operate as a computer program working on a colossal lookup table, and the answer can be summarized by Chaitin's dictum ``compression is comprehension''~\cite{chaitina}. We believe in fact, just as Bennett does~\cite{bennett}, that decompression, i.e. the calculation to arrive at the decompressed data, is also a key element, but one that falls into the algorithmic realm that we are defending.


One may have the feeling that Searle's point was related to grounding and semantics versus syntax, and thereby that algorithms are still of syntactic nature, but Searle recognizes that he is not claiming anything metaphysical. What we are claiming is that it follows from the impossibility of certain computer programs to explain understanding of the human mind is that his human understanding must therefore be related to features of highly algorithmic nature (e.g. compression/decompression) and thus that not all computer programs are the same, particularly when resources are involved. And this constraint of resources comes from the brain optimization of all sorts of cost functions achieved by striving for optimal learning~\cite{entropy} such as minimizing energy consumption, to mention but one example, and in this optimal behavior \textit{algorithmic probability}~\cite{solomonoff} must have an essential role.

In light of the theoretical triviality of passing the Turing test, it has been stressed~\cite{parberry,aaronson} the need to consider the question of resources and therefore of \textit{computational complexity}. This means that the mind harness mechanisms to compress large amounts of information in efficient ways. An interesting connection to \textit{integrated information} can be established by way of compression. Imagine one is given two files. One is uncompressible and therefore any change to the uncompressed file can be simply replicated in the compressed file, the files being effectively the same. This is because any given bit is independent of every other in both versions, compressed and uncompressed. But reproducing any change in a compressible file leading to an effect in an uncompressed file requires a cryptic change in the former,  because the compression algorithm takes advantage of short and long-range regularities, i.e. non-independency of the bits, encoded in a region of the compressed file that may be very far afield in the uncompressed version. This means that uncompressible files have little integrated information but compressible files have a greater degree of integrated information. Similar ideas are explored in~\cite{maguire}, making a case against integrated information theory by arguing that it does not conform to the intuitive concept of consciousness. For what people mean by the use of `consciousness' is that a system cannot be broken down~\cite{tononi}; if it could it would not amount to ``consciousness''.

In practice, search engines and the \textit{web} constitute a sort of lookup table of exactly the type attacked by Searle's CRA. Indeed, the probability of finding a webpage containing a permutation of words representing a short sentence is high and tends to increase over time, even though the probability grows exponentially slowly due to the combinatorial explosion. The web contains about 4.73 billion pages (\url{http://www.worldwidewebsize.com/} as estimated and accessed on Wednesday, 07 January, 2015) with text mostly written by humans. But that the web seen through a search engine is a lookup table of sorts is an argument against the idea that the web is a sort of  ``global mind'' and therefore consonant with Searle's argument. Indeed, there is very little evidence that anything in the web is in compressed form (this is slightly different from the Internet in general, where many transactions and protocols implement compression in one way or another, e.g. encryption). We are therefore proposing that compression is a necessary condition for minds and consciousness, but clearly not a sufficient one (cf. compressed files, such as PNG images in a computer hard drive).

It has been found in field experiments that animals display a strong behavioral bias. For example, in Reznikova's communication experiments with ants, simpler instructions were communicated faster and more effectively than complex ones by scout ants searching for patches of food randomly located in a maze. The sequences consist of right and left movements encoded in binary (R and L) (see \url{http://grahamshawcross.com/2014/07/26/counting-ants/} Accessed on Dec 23, 2014.)~\cite{reznikova,reznikova2,reznikova3} that are communicated by the scout ants returning to their foraging team in the colony to communicate instructions for reaching the food. We have quantified some of these animal behavior experiments confirming the author's suspicion that algorithmic information theory could account for the biases with positive results~\cite{zenilmarshalltegner}.

Humans too display biases in the same \textit{algorithmic} direction, from their motion trajectories~\cite{peng} to their perception of reality~\cite{chater}. Indeed, we have shown that cognition, including visual perception and the generation of subjective randomness, shows a bias that can be accounted for with the seminal concept of algorithmic probability~\cite{gauvrit2014a,gauvrit2014b,gauvrit2014c,kempe,Mathy2014}. Using a computer to look at human behavior in a novel fashion, specifically by using a reverse Turing test where what is assessed is the human mind and an ``average'' Turing machine or computer program implementing any possible compression algorithm, we will show that the human mind behaves more like a machine. We will in effect reverse the original question Turing posed via his imitation game as to whether machines behave like us.

\section{Algorithmic complexity as model of the mind}

Since the emergence of the Bayesian paradigm in cognitive science, researchers have expressed the need for a formal account of complexity based on a sound complexity measure. They have also struggled to find a way of giving a formal normative account of the probability that a deterministic algorithm produces a given sequence, such as the heads-or-tails string ``HTHTHTHT'', which intuitively looks like the result of a deterministic process even if it has the same probability of occurrence as any other sequence of the same length according to classical probability theory, which assigns a probability of $1/2^8$ to all sequences of size 8.

\subsection{From bias to Bayes}

Among the diverse areas of cognitive science that have expressed a need for a new complexity measure, the most obvious is the field of probabilistic reasoning. The famous work of Kahneman, Slovic and Tversky (1982) aimed at understanding how people reason and make decisions in the face of uncertain and noisy information sources. They showed that humans were prone to many errors about randomness and probability. For instance, people tend to claim that the sequence of heads or tails ``HTTHTHHHTT'' is more likely to appear when a coin is tossed than the series ``HHHHHTTTTT''.

In the ``heuristics and bias'' approach advocated by Kahneman and Tversky, these ``systematic'' errors were interpreted as biases inhering in human psychology, or else as the result of using faulty heuristics. For instance, it was believed that people tend to say that ``HHHHHTTTTT'' is less random than ``HTTHTHHHTT'' because they are influenced by a so-called representativeness heuristic, according to which a sequence is more random the better it conforms to prototypical examples of random sequences. Human reasoning, it has been argued, works like a faulty computer. Although many papers have been published about these biases, not much is known about their causes.

Another example of a widespread error is the so-called ``equiprobability bias'' \cite{lecoutre}, a tendency to believe that any random variable should be uniform (with equal probability for every possible outcome). In the same vein as the seminal investigations by Kahneman et al.~\cite{kahneman}, this bias too, viz.the erroneous assumption that randomness implies uniformity, has long been interpreted as a fundamental flaw of the human mind.

During the last decades, a paradigm shift has occurred in cognitive science. The ``new paradigm''---or Bayesian approach---suggests that the human (or animal) mind is not a faulty machine, but a probabilistic machine of a certain type. According to this understanding of human cognition, we all estimate and constantly revise probabilities of events in the world, taking into account any new pieces of information, and more or less following probabilistic (including Bayesian) rules.

Studies along these lines often try to explain our probabilistic errors in terms of a sound intuition about randomness or probability applied in an inappropriate context. For instance, a mathematical and psychological reanalysis of the equiprobability bias was recently published~\cite{gauvrit2014}. The mathematical theory of randomness, based on  algorithmic complexity (or on entropy, as it happens) does in fact imply uniformity. Thus, claiming that the intuition that randomness implies uniformity is a bias does not fit with mathematical theory. On the other hand, if one follows the mathematical theory of randomness, one must admit that a combination of random events is, in general, not random anymore. Thus, the equiprobability bias (which indeed is a bias, since it yields frequent faulty answers in the probability class) is not, we argue, the result of a misconception regarding randomness, but a consequence of the incorrect intuition that random events can be combined without affecting their property of randomness.

We now believe that when we have to compare the probability that a fair coin produces either ``HHHHHTTTTT'' or any other 10-item long series, we do not really do so. One reason is that the question is unnatural: our brain is built to estimate the probabilities of the causes of observed events, not the \textit{a priori} probability of such events. Therefore, say researchers, when we have participants rate the probability of the string $s=$``HHHHHTTTTT'' for instance (or any other), they do not actually estimate the probability that such a string will appear on tossing a fair coin, which we could write as $P(s|R)$ where $R$ stands for ``random process'', but the reverse probability $P(R|s)$, that is, the probability that the coin is fair (or that the string is genuinely random), given that it produced $s$. Such a probability can be estimated using Bayes' theorem:

$$P(R|s)= \frac{P(s|R)P(R)}{P(s|R)P(R)+P(s|D)P(D)},$$

\noindent where $D$ stands for ``not random'' (or deterministic).

In this equation, the only problematic element is $P(s|D)$, the probability that an undetermined but deterministic algorithmic will produce $s$. It was long believed that no normative measure of this value could be assumed, although some authors had the intuition that is was linked to the complexity of $s$: simple strings are more likely to appear as a result of an algorithm than complex ones.

The algorithmic theory of information actually provides a formal framework for this intuition. The algorithmic probability of a string $s$ is the probability that a randomly chosen program running on a Universal (prefix-free) Turing Machine will produce $s$ and then halt. It therefore serves as a natural formal definition of $P(s|D)$. As we will see below, the algorithmic probability of a string $s$ is inversely linked to its (Kolmogorov-Chaitin) algorithmic complexity, defined as the length of the shortest program that produces $s$ and then halts: simpler strings have a higher algorithmic probability.

One important drawback of algorithmic complexity is that it is not computable. However, there now exist methods to approximate the probability, and thus the complexity, of any string, even short ones (see below), giving rise to a renewed interest in complexity in the cognitive sciences.

Using these methods~\cite{gauvrit2014b}, we can compute that, with a prior of 0.5, the probability that the string HHHHHTTTTT is random amounts to 0.58, whereas the probability that HTTHTHHHTT is random amounts to 0.83, thus confirming the common intuition that the latter is ``more random'' than the former.

\subsection{The Coding theorem method}

One method for assessing Kolmogorov-Chaitin complexity, namely lossless compression, as epitomized by the Lempel-Ziv algorithm, has been long and widely used. Such a tool, together with classical Shannon entropy~\cite{wang}, has been used recently in neuropsychology to investigate the complexity of EEG or fMRI data~\cite{maguire} and~\cite{adenauer}. Indeed, the size of a compressed file gives an indication of its algorithmic complexity. The size of a compressed file is, in fact, an upper bound of true algorithmic complexity. On the one hand, compression methods have a basic flaw; they can only recognize statistical regularities and are therefore implementations of variations of entropic measures, only assessing the rate of entropic convergence based on repetitions of strings of fixed sliding-window size. If lossless compression algorithms work to approximate Kolmogorov complexity, it is only because compression is a sufficient test for non-randomness, but they clearly fail in the other direction, when it is a matter of ascertaining whether something is the result of or is produced by an algorithmic process (e.g. the digits of the mathematical constant $\pi$). That is, they cannot find structure that takes other forms than simple repetition. On the other hand, compression methods are inappropriate for short strings (of, say, less than a few hundreds symbols). For short strings, lossless compression algorithms often yield files that are longer than the strings themselves, hence providing results that are very unstable and difficult, if not impossible, to interpret.

In cognitive and behavioral science, researchers usually deal with short strings of at most a few tens of symbols, for which compression methods are thus useless. This is the reason they have long relied on tailor-made measures instead.

The Coding theorem method~\cite{delahaye,solerplos} (CTM) has been specifically designed to address this challenge. By using CTM, researchers have provided values for the ``algorithmic complexity for short strings'' (which we will abbreviate as ``$ACSS$''). $ACSS$ is available freely as an R-package (named $ACSS$), or through an online (\url{www.complexitycalculator.com} Accessed on 26 Dec, 2014) complexity calculator~\cite{gauvrit2014a}.

At the root of $ACSS$ is the idea that algorithmic probability may be used to capture algorithmic complexity. The algorithmic probability of a string $s$ is defined as the probability that a universal prefix-free Turing machine $U$ will produce $s$ and then halt. Formally,

$$m(s)=\sum_{U(p)=s} 1/2^{-|p|}$$

The algorithmic complexity~\cite{kolmogorov,chaitina} of a string $s$ is defined as the length of the shortest program that, running on a universal prefix-free~\cite{levin} Turing machine $U$, will produce $s$ and then halt. Formally,

$$K(s) = \min\{|p| : U(p) = s\}$$

$K_U(s)$ and $m_U(s)$ both depend on the choice of the Turing machine $U$. Thus, the expression ``the algorithmic complexity of $s$'' is, in itself, a shortcut. For long strings, this dependency is relatively small. Indeed, the \textit{invariance theorem} \cite{solomonoff,kolmogorov,chaitina} states that for any $U$ and $U^\prime$, two universal prefix-free Turing machines, there exists a constant $c$ independent of $s$ such that

$$|K_U(s) - K_{U^\prime}(s)| < c$$

The constant $c$ can be arbitrarily large. If one wishes to approximate the algorithmic complexity of short strings, the choice of $U$ is thus determinant.

To overcome this inconvenience, we can take advantage of a formal link established between algorithmic probability and algorithmic complexity. The \textit{algorithmic Coding theorem}~\cite{levin} states that

$$K_U(s) = - \log_2(m_U(s)) + O(1)$$

This theorem can be used to approximate $K_U(s)$ through an estimation of $m_U(s)$.

Instead of choosing a particular arbitrary universal Turing machine and feeding it with random programs, Delahaye and Zenil (2012) had the idea (equivalent in a formal way) of using a huge sample of Turing machines running on blank tapes. By doing so, they built a ``natural'' (a quasi-lexicographical enumeration) experimental distribution approaching $m(s)$, conceived of as an average $m_U(s)$.

They then defined $ACSS$(s) as $-\log_2(m(s)$). $ACSS$(s) approximates an average $K_U(s)$. To validate the method, we studied how $ACSS$ varies under different conditions. It has been found that $ACSS$ as computed with different huge samples of small Turing machines remained stable~\cite{gauvrit2014a}. Also, several descriptions of Turing machines did not alter $ACSS$~\cite{zenilalgo}. Using cellular automata instead of Turing machines, they showed that $ACSS$ remained relatively stable. On a more practical level, $ACSS$ is also validated by experimental results. For instance, as we will see below, $ACSS$ is linked to human complexity perception. And it has found applications in graph theory and network biology~\cite{zenilgraph}.

\subsection{The Block Decomposition Method}

$ACSS$ provides researchers with a user-friendly tool~\cite{gauvrit2014b} for assessing the algorithmic complexity of very short strings. Despite the huge sample (billions) of Turing machines used to build $m(s)$ and $K(s)$ using the Coding theorem method, two strings of length 12 were never produced, yet $ACSS$ can safely be used for strings up to length 12 by assigning the missing two the greatest complexity in the set plus 1.

The same method used with two-dimensional Turing machines produced a sample of binary patterns on grids, but here again of limited range. Not every $5\times 5$ grid is present in the output but one can approximate any square grid $n\times n$ by decomposing it to $4\times 4$ squares for which there is a known estimation approximated by a two-dimensional Coding theorem method. The method simply involved running Turing machines on lattices rather than tapes, and then all the theory around the Coding theorem could be applied and a two-dimensional Coding theorem method conceived.

This idea of decomposing led to filling in the gap between the scope of classical lossless compression methods (large strings) and the scope of $ACSS$ (short strings). To this end, a new method based on $ACSS$ was developed: the Block decomposition method~\cite{zenilgraph}.

The basic idea at the root of the Block decomposition method is to decompose strings for which we have exact approximations of their Kolmogorov complexity into shorter substrings (with possible overlapping). For instance, the string ``123123456'' can be decomposed into 6-symbol subsequences with a 3-symbol overlap ``123123'' and ``123456''. Then, the block decomposition method takes advantage of known information-theoretic properties by penalizing $n$ repetitions that can be transmitted by using only $\log_2(n)$ bits through the formula:

$$C(s) = \sum_p\log_2(n_p) + K(p)$$

\noindent where $p$ denotes the different types of substrings, $n_p$ the number of occurrences of each $p$, and $K(p)$ the complexity of $p$ as approximated by $ACSS$. As the formula shows, the Block decomposition method takes into account both the local complexity of the string and the (possibly long distance) repetitions.

With the Coding theorem method leading to $ACSS$, the block decomposition method and compression algorithms, the whole range of string lengths can now be covered with a family of formal methods. Within the field of cognitive and behavioral science, these new tools are really relevant. In the next section, we will describe three recent areas of cognitive science in which algorithmic complexity plays an important role. In these areas $ACSS$ and the Block decomposition method have been key in the approximation of the algorithmic complexity of short (2-12) and medium (10-100) length strings and arrays.

\section{Cognition and complexity}

An early example of an application to cognition of the Coding theorem and the Block decomposition methods was able to quantify the short sequence complexity in Reznikova's ant communication experiments, validating their results~\cite{zenilbehavior}. Indeed, it was found that ants communicate simpler strings (related to instructions for getting to food in a maze) in a shorter time, thus establishing an experimental connection between animal behavior, algorithmic complexity and time complexity.

The idea was taken further to establish a relation between information content and energy (spent both in foraging and communicating) to establish a fitness landscape based on the thresholds between these currencies as reported in~\cite{entropy}. As shown in~\cite{entropy}, if the environment is too predictable the cost of information-processing is very low. In contrast, if the environment is random, the cost is at a maximum. The results obtained by these information-processing methods suggest that organisms with better learning capabilities save more energy and therefore have an evolutionary advantage.

In many ways, animal behavior (including, notably, human behavior) suggests that the brain acts as a data compression device. For example, despite our very limited memory, it is clear we can retain long strings if they have low algorithmic complexity (e.g. the sequence 123456$\ldots$ vs the digits of $\pi$, see below). Cognitive and behavioral science  deals for the most part with small series, barely exceeding a few tens of values. For such short sequences, estimating the algorithmic complexity is a challenge. Indeed, until now, behavioral science has largely relied on a subjective and intuitive account of complexity (e.g. Reznikova's ant communication experiments). Through the ``Coding theorem method'', however, it is now possible to obtain a reliable approximation of the algorithmic complexity of any string length~\cite{gauvrit2014a} and the methods have been put to the test and applied to validate several intuitions put forward by prominent researchers in cognitive science over the last few decades. We will describe some of these intuitions and explain how they have been confirmed by experimental studies.

\subsection{Working memory}

Researchers in psychology have identified different types of memory in humans and other animals. Two related kinds of memory have been particularly well studied: short-term memory and working memory. Short-term memory refers to a kind of cognitive resource that allows the storage of information over a few seconds or minutes. According to a recent account of short-term memory~\cite{barrouillet2}, items stored in short-term memory quickly and automatically decay as times passes, as a mere result of the passage of time, unless attention is devoted to the reactivation of the to-be-remembered item. Thus, constant reactivation is necessary to keep information stored for minutes. This is done by means of an internal language loop (one can continuously repeat the to-be-memorized items) or by using mental imagery (one can produce mental images of the items). On the other hand, some researchers believe that the decay observed in memory is entirely due to interference by new information~\cite{oberauer}. Researchers thus disagree as to the reasons while they all acknowledge that there is indeed a quick decay of memory traces. The span of short-term memory is estimated at around 7 items~\cite{miller,jesper}. To arrive at this estimation, psychologists use simple memory span tasks in which participants are required to repeat series of items of increasing length. For instance, the experimenter says ``3, 5'' and the participants must repeat ``3, 5''. Then the experimenter says ``6, 2, 9'' out loud, which is repeated by the participant. Short-term memory span is defined as the length of the longest series one can recall. Strikingly, this span is barely dependent on the type of item to be memorized (letters, digits, words, etc.).

Two observations, however, shed new light on the concept of short-term memory. The first is the detrimental effect of cognitive load on short-term memory span. For instance, if humans usually recall up to 7 items in a basic situation as described above, the number of items correctly recalled drops if an individual must perform another task at the same time. For instance, if a subject had to repeat ``baba'' while memorizing the digits, s/he would probably not store as many as 7 items~\cite{barrouillet}. More demanding interfering tasks (such as checking whether equations are true or generating random letters) lead to even lower spans. The second observation is that when there is a structure in the to-be-recalled list, one can retain more items. For instance, it is unlikely that anyone can memorize, in a one-shot experiment, a series such as ``3, 5, 4, 8, 2, 1, 1, 9, 4, 5''. However, longer series such as ``1, 2, 3, 4, 5, 6, 5, 4, 3, 2, 1'' will be easily memorized and recalled.

The first observation led to the notion of working memory. Working memory is a hypothetical cognitive resource used for both storing and processing information. When participants must recall series while performing a dual task, they assign part of their working memory to the dual task, which reduces the part of working memory allocated to storage, i.e. short-term memory~\cite{baddeley}.

The second observation led to the notion of chunking. It is now believed that humans can divide to-be-recalled lists into simple sub-lists. For instance, when they have to store ``1, 2, 3, 4, 5, 3, 3, 3, 1, 1, 1'' they can build three ``chunks'': ``1, 2, 3, 4, 5'', ``3, 3, 3'' and ``1, 1, 1''. As this example illustrates, chunks are not arbitrary: they are conceived to minimize complexity, and thus act as compression algorithms~\cite{mathy}. An objective factor showing that chunking does occur is that while recalling the above series, people make longer pauses between the hypothetical chunks than within them. Taking these new concepts into account, it is now believed that the short-term memory span is not roughly 7 items, as established using unstructured lists, but more accurately around 4 chunks~\cite{cowan}.

The short-term memory span is thus dependent on the type of list to be memorized. With structured series, one can recall more items, but fewer chunks. This apparent contradiction can be overcome by challenging the clear-cut opposition between short-term memory span and working-memory span. To overcome the limitations of short-term memory, humans can take advantage of the structure present in some series. Doing so requires using the ``processing'' component of working memory to analyze the data~\cite{mathy}. Thus, even simple tasks where subjects must only store and recall series of items do tax the processing part of working memory.

According to this hypothesis, working memory works as a compression algorithm, where part of the resource is allocated to data storage while another part is dedicated to the ``compression/decompression'' program. Recent studies are perfectly in line with this idea. Mathy et al. (2014) used a Simon$\textregistered$, a popular 80s game in which one must reproduce series of colors (chosen from among a set of 4 colors) of increasing length, echoing classical short-term memory tasks. They show that the algorithmic complexity of a series is a better predictor of correct recall than the length of the string. Moreover, when participants make mistakes, they generate, on average, an (incorrect) string of lower algorithmic complexity than the string to be remembered.

All these experimental results suggest that working memory acts as a compression algorithm. In normal conditions, compression is lossless, but when the items to be remembered exceed working memory capacity, it may be that lossy compression is used instead.

\subsection{Randomness perception}

Humans share some intuitive and seemingly innate (or at least precocious) concepts concerning mathematical facts, biology, physics and na{\"i}ve psychology. This core knowledge~\cite{spelke} is found in babies as young as a few months. In the field of na{\"i}ve mathematics, for instance, it has been shown that people share some basic knowledge or intuitions about numbers and geometry. Humans and other animals both manifest the ability to discriminate quantities based on an ``Approximate Number Sense'', thanks to which one can immediately feel that 3 objects are more than 2, or 12 than 6~\cite{dehaene}. Because this number sense is only approximate, it doesn't allow us to discriminate 12 from 15, for instance. There are many indications that we do have an innate sense of quantity. For instance, both children and adults are able to ``count'' quantities not greater than 4 almost immediately, whereas larger quantities require a counting strategy. When faced with 1, 2 or 3 objects, we immediately perceive how many there are without counting. This phenomenon is known as ``subitizing''~\cite{mandler}. Babies are young as six months~\cite{xu}, apes~\cite{boysen} and even birds~\cite{pepperberg} are able to perform basic comparisons between small numbers. However, for quantities above 4, we are bound to resort to learned strategies like counting.

The same results have been found in the field of na{\"i}ve geometry. In a study of an Amazonian indigenous group, Dehaene et al (2006) found that people without any formal mathematical culture naturally reason in terms of points and lines. They also perceive symmetry and share intuitive notions of topology, such as connexity or simple connexity (holes).

Recent findings in the psychology of subjective randomness suggest that probabilistic knowledge may well figure on the list of core human knowledges. T\'egl\'as et al (2011) exposed 1-year old infants to images of jars containing various proportions of colored balls. When a ball is randomly taken from the jar, the duration of the infant's gaze is recorded. Also, one can easily compute the probability of the observed event. For instance, if a white ball is chosen, the corresponding probability is the proportion of white balls in the jar. It turned out that gazing time is correlated to the probability of the observed event: the more probable the event, the shorter the gaze. This is interpreted as proof that 1-year old children already have an intuitive and approximate probability sense, in the same way they have an approximate number sense.

This probability sense is even more complex and rich than the previous example suggests. Xu and Garcia (2008) proved that infants even younger (8 months old) could exhibit behaviors showing a basic intuition of Bayesian probability. In their experiment, children saw an experimenter taking white or red balls out of an opaque jar. The resulting sample could exhibit an excess of red or white, depending on the trial. After the sample was removed, the jar was uncovered, revealing either a large proportion of red balls or a large proportion of white balls. It was observed that children stared longer at the scene if what transpired was less probable according to a Bayesian account of sampling. For instance, if the sample had a large majority of red balls but the jar, once uncovered, showed an excess of white ones, the infant's gaze would be trained on the scene  longer.

In the same vein, Ma and Xu (2009) presented 9 and 10-month old children with samples from a jar containing as many yellow as red balls. However, the sampling could be done either by a human or a robot. Ma and Xu show that children expect any regularity to come from a human hand. This experiment was meant to trace our intuition of intentionality behind any regularity. However, it also shows that humans have a precocious intuition of regularity and randomness.

Adults are able to discriminate more finely between strings according to their complexity. For example, using classical randomness perception tasks, Matthews (2013) had people decide whether binary strings of length 21 were more likely to have been produced by a random process or a deterministic one. Matthews was primarily interested in the contextual effects associated with such a task. However, he also shows that adults usually agree about which strings are more random. A re-analysis of Mathews' data~\cite{gauvrit2014b} showed that algorithmic complexity is actually a good predictor of subjective randomness---that is, the probability that a human would consider that the string looks random.

In attempting to uncover the sources of our perception of randomness, Hsu et al (2010) have suggested that it could be learned from the world. More precisely, according to the authors, our two-dimensional randomness approach~\cite{kolmo2d} could be based on the probability that a given pattern appears in visual natural scenes. They presented a series of patterns, which were $4\times 4$ grids with 8 black cells, and 8 white cells. Subjects had to decide whether these grids were ``random'' or not. The proportion of subjects answering ``random'' was used as a measure of subjective randomness. The authors also scanned a series of still nature shots. The pictures were binarized to black and white using the median as threshold and every $4\times 4$ grid was extracted from them. From this dataset, a distribution was computed. Hsu et al (2010) found that the probability that a grid appears in random photographs of natural scenes was correlated to the human estimate: the more frequent the grid, the less random it is rated. The authors interpret these results as evidence that we learn to perceive randomness through our eyes. An extension of the Hsu et al study (2010)  confirmed the correlation, and found that both subjective randomness and frequency in natural scenes were correlated to the algorithmic complexity of the patterns~\cite{gauvrit2014c}. It was found that natural scene statistics explain in part how we perceive randomness/complexity.

In contradistinction to the aforementioned experiments according to which even children under one year displayed the ability to  detect regularities, these results suggest that the perception of randomness is not innate. Our sense of randomness could deploy very early in life, based on visual scenes we see and an innate sense of statistics, but it evolves over time.

Our results in~\cite{gauvrit2014c}, suggest that the mind is intrinsically wired to believe that the world is algorithmic in nature, that what happens in it is likely the output of a random computer program and not of a process producing uniform classical randomness. To know if this is the result of biological design or a developed ``skill'' one can look at whether and how people develop this view during lifetime. Preliminary results suggest that this algorithmic view or filter about the world we are equipped with is constructed over time reaching a peak of algorithmic randomness production and perception at about age 25 years (see~\url{http://complexitycalculator.com/hrng/}). This means that the mind adapts and learns from experience and for some reason it develops the algorithmic perception before changing again. We would not have developed such a worldview peaking at reproductive age had it no evolutionary advantage as it seems unlikely to be a a neutral feature of the mind about the world, as it affects the way it perceives events and therefore how it learns and behaves. And this is not exclusive to the human mind but to animal behavior as shown in~\cite{zenilmarshalltegner}. All this evidence points out towards the same direction, that the world is or appears to us highly algorithmic in nature, at least transitionally.

\subsection{Culture and structure}

In different areas of social science and cognitive science, it has been found that regularities may arise from the transmission of cultural items~\cite{smith}. For instance, in the study of rumors, it has been found that some categories of rumor disseminate easily, while others simply disappear. The ``memes'' that are easily remembered, believed and transmitted share some basic properties, such as a balance between expected and unexpected elements, as well as simplicity~\cite{bronner}. Too-complex stories, it is believed, are too hard to remember and are doomed to disappear. Moreover, in spreading rumors the transmitters make mistakes, in a partially predictable way. As a rule, errors make the message simpler and thus easier to retain.

For instance, Barrett and Nihoff (2001) had participants learn fabricated stories including some intuitive and some counter-intuitive elements. They found that slightly counterintuitive elements were better recalled than intuitive elements. In research along the same lines, Atran and Norenzayan (2004) found an inverse U-shaped curve between the proportion of counter-intuitive elements and the correct recall of stories.

Two complementary concepts are prominent in the experimental psychology of cultural transmission. \textit{Learnability} measures how easily an item (story, word, pattern) is remembered by someone. More learnable elements are more readily disseminated. They are selected in a ``darwinian'' manner, since learnability is a measure of how adapted the item is to the human community in which it will live or die.

\textit{Complexity} is another important property of cultural items. In general, complex items are less learnable, but there are exceptions. For instance, a story without any interest will not be learned, even if highly simple. Interest, unexpectedness or humoristic level also play a role in determining how learnable and also how ``buzzy'' an item is.

When researchers study complex cultural items such as stories, arguments, or even theories or religions, they are bound to measure complexity and learnability by using more or less tailor-made tools. Typically, researchers use the number of elements in a story, as well as their interrelations to rate its complexity. The number of elements and relations someone usually retains is an index of learnability.

The paradigm of iterative learning is an interesting way to study cultural transmission and investigate how it yields to structural transformations of the items to be transmitted. In the iterative learning design, a first participant has to learn a message (a story, a picture, a pseudo-language, etc.) and then reproduce or describe it to a second participant. The second participant will then describe the message to a third, etc. A chain of transmission can comprise tens of participants.

Certainly, algorithmic complexity is not always usable in such situations. However, in fundamental research in cognitive psychology, it is appropriate. Instead of investigating complex cultural elements such as rumors or pseudo-sciences, we turned to simpler items, in the hope of achieving a better understanding of how structure emerges in language. Some pseudo-languages have been used~\cite{kirby} and showed a decrease in intuitive complexity along the transmission chains.

In a more recent study, researchers used two-dimensional patterns of points~\cite{kempe}. Twelve tokens were randomly placed on $10\times10$ grids. The first participant had to memorize the pattern in 10 seconds, and then reproduce it on a blank grid, using new tokens. The second participant then had to reproduce the first participant's response, and so on. Participants in a chain were sometimes all adults, and sometimes all children. Learnability can be defined as the number of correctly placed tokens. As expected, learnability continuously increases within each transmission chain, at the same rate for children and adults. As a consequence of increasing learnability, each chain converges toward a highly learnable pattern (which also is simple), different from one chain to the other.

The fact that children's and adults' chains share the same rate of increase in learnability is striking, since they have different cognitive abilities$\ldots$, but may be explained by complexity. Indeed, the algorithmic complexity of the patterns within each chain decreases continuously, but at a faster rate in children's chains than adult chains.

The Less-is-More effect~\cite{kersten} in psycholinguistics designates the fact that babies, with low cognitive abilities, are far better than adults at learning language. Thus, less cognitive ability could actually be better when it comes to learning language. One possible explanation of the paradox relies on complexity: as a consequence of the reduced cognitive power of babies, they could be bound to ``compress the data'' more---build a simpler though possibly faulty representation of the available data they have access to. The recent study of Kempe et al. (2015) supports this idea (albeit remotely), by showing that children achieve the same increase in learnability as adults by a much quicker decrease in algorithmic complexity, which can be a consequence of reduced cognitive power.

\section{Concluding remarks}

We have surveyed most of what has formally been done to connect and explain cognition through computation. From Turing to Searle to Tononi to our own work. It may seem that a mathematical/computational framework is taken too far, for example in interpreting how humans may remember long sequences and that remembering such sequences may probably never be useful for any organism. But it is not the object in question (string) at the core of the argument but the mechanism. Learning by systematic incorporation of information from the environment is what ultimately a DNA sequence is, and what the evolution of brains ultimately allowed to speed up the understanding of an organism's environment. Of course, the theoretical implications of a Turing machine model---such as halting, having an unbounded tape etc.---are only irrelevant if taken trivially as an analogy for biological processes, but this is missing the point. The point is not that organisms or biological processes are or may look like Turing machines, but that Turing machines are mechanistic explanations of behavior, and as shown by algorithmic probability, they are an optimal model for hypothesis testing of, in our opinion, the uttermost relevance for biological reality and what we call the algorithmic cognition approach to cognitive sciences at the forefront of pattern recognition, and multisensory integration.

It may also be believed that computational and algorithmic complexity focus only on classifying algorithmic problems according to their inherent difficulty or randomness and as such that it cannot be made equal to models of cognition. Moreover, that algorithmic analysis can give meaningful results only for a relatively narrow class of algorithms that on one hand are not too general and on the other hand are not too complex and detailed and reflect the main ingredients of cognitive processes. These beliefs would only hold if the power and universality of the results of the field of algorithmic probability (the other side of the same coin of algorithmic complexity) as introduced by Solomonoff~\cite{solomonoff} and Levin~\cite{levin} were not understood, an introduction giving it proper credit, as the optimal theory of formal induction/inference, and therefore learning, can be found in~\cite{miracle}. The fact that the theory is based on the Turing machine model is irrelevant and cannot be used as an objection as shown, for example, by the Invariance Theorem~\cite{solomonoff,levin}, which not only also applies to algorithmic probability but Levin showed that disregarding the computational model the algorithmic probability converges and the only universal semi-measure dominates~\cite{miracle}. We have adopted this powerful theory and used it to advance the algorithmic cognition approach we have devised specific tools with a wide range of applications~\cite{zenilchaos,zenilca} that also conform to animal and human  behavior~\cite{zenilbehavior,gauvrit2014a,gauvrit2014b,gauvrit2014c,kempe,Mathy2014,zenilmarshalltegner}.

Just as Tononi et al. made substantial progress in discussing an otherwise more difficult topic by connecting the concept of consciousness to information theory. We have offered what we think is an essential and what appears a necessary connection between the concept of cognition and algorithmic information theory. Indeed, within cognitive science, the study of working memory, of probabilistic reasoning, the emergence of structure in language, strategic response, and navigational behavior is cutting-edge research. In all these areas we have made contributions~\cite{zenilbehavior,gauvrit2014a,gauvrit2014b,gauvrit2014c,kempe,Mathy2014,zenilmarshalltegner} based upon algorithmic complexity as a useful normative tool, shedding light on mechanisms of cognitive processes. We have proceeded by taking the mind as a form of algorithmic compressor, for the reasons provided in this survey emerging from the simple fact that the mind operates otherwise than through a lookup table and the ways in which the mind manifests biases that seem not to be accounted for but by algorithmic probability based purely on computability theory. The research promises to generate even more insights into the fundamentals of human and animal cognition~\cite{zenilmarshalltegner}, with cross-fertilization taking place from the artificial to the natural realms and viceversa, as algorithmic measures of the properties characterizing human and animal behavior can be of use in artificial systems like robotics~\cite{zenilrobot}, which amounts to a sort of reverse Turing test as described in~\cite{zenilphilo} and~\cite{zeniljetai}.

\subsection*{Acknowledgments}

The authors are indebted to the anonymous referees and to the hard work of the members of the Algorithmic Nature Group, LABORES\\(\url{http://www.algorithmicnaturelab.org}).


\begin{thebibliography}{10}

\bibitem[Aaronson 2013]{aaronson} Aaronson S (2013), Why Philosophers Should Care About Computational Complexity. In Copeland BJ, Posy C and Shagrir O (Eds), Computability: Turing, G{\"o}del, Church, and Beyond, MIT Press, 2013, pp
261--328.
\bibitem[Atran \& Norenzayan 2004]{atran} Atran S, Norenzayan A (2004) Religion's evolutionary landscape: Counterintuition, commitment, compassion, communion. \emph{Behav Brain Sci} 27:713-770
\bibitem[Baddeley 1992]{baddeley} Baddeley A (1992) Working memory. \emph{Science} 255(5044):556-559
\bibitem[Barrett \& Nyhof 2001]{barrett} Barrett J L, Nyhof M A (2001) Spreading non-natural concepts: The role of intuitive conceptual structures in memory and transmission of cultural materials. \emph{J Cogn \& Culture}. 1(1):69-100
\bibitem[Barrouillet et al. 2004]{barrouillet} Barrouillet P, Bernardin S, Camos V (2004) Time constraints and resource sharing in adults' working memory spans. \emph{J Exp Psychol Gen} 133(1):83
\bibitem[Barrouillet et al. 2009]{barrouillet2} Barrouillet P, Gavens N, Vergauwe E, et al (2009) Working memory span development: a time-based resource-sharing model account. \emph{Dev Psychol} 45(2):477
\bibitem[Bennett 1998]{bennett} Bennett CH (1988) Logical depth and physical complexity. In Herken R (ed.) \textit{The Universal Turing Machine. A Half-Century Survey.} pp. 227--257, Oxford: Oxford University Press.
\bibitem[Boysen \& Hallberg 2000]{boysen} Boysen ST, Hallberg KI (2000) Primate numerical competence: contributions toward understanding nonhuman cognition. \emph{Cog Science} 24(3):423-443.
\bibitem[Brenner 2012]{brenner} Brenner S (2012). Turing centenary: Life's code script, \emph{Nature,} 482, 461.
\bibitem[Bronner 2010]{bronner} Bronner G (2010). Le succ\`es d'une croyance. \emph{Ann Soc} 60(1):137-160.
\bibitem[Casali et al. 2013]{adenauer} Casali AG, Gosseries O, Rosanova M, Boly M, Sarasso S, Casali KR, Casarotto S, Bruno M-A, Laureys S, Tononi G, Massimini M, (2013) A Theoretically Based Index of Consciousness Independent of Sensory Processing and Behaviour, vol. 5(198), \emph{Sci. Transl. Med.}.
\bibitem[Chaitin 1966]{chaitina} Chaitin GJ On the length of programs for computing finite binary sequences. \textit{J. ACM} 13(4), 547--569. 
\bibitem[Chaitin 2006]{chaitin} Chaitin G, The Limits of Reason, \emph{Scientific American} 294, No. 3, pp. 74--81, March 2006.
\bibitem[Chater 1999]{chater} Chater N The search for simplicity: A fundamental cognitive principle? \emph{The Quarterly Journal of Experimental Psychology,} 52 (A), 273-302, 1999.
\bibitem[Church \& Rosser 1936]{church} Church A, Rosser JB, (1936), Some properties of conversion, \emph{Transactions of the American Mathematical Society,} 39 (3): 472--482.
\bibitem[Cowan 2001]{cowan} Cowan, N. (2001) The magical number 4 in short-term memory: a reconsideration of mental storage capacity. \emph{Behav Brain Sci} 24(1):87?114.
\bibitem[Dehaene 2011]{dehaene} Dehaene S (2011) \emph{The number sense: How the mind creates mathematics.} Oxford University Press, Oxford.
\bibitem[Dehaene et al. 2006]{dehaene2} Dehaene S, Izard V, Pica P, Spelke E (2006) Core knowledge of geometry in an Amazonian indigene group. \emph{Science} 311(5759):381-384
\bibitem[Delahaye \& Zenil 2012]{delahaye} Delahaye J-P, Zenil H (2012) Numerical evaluation of algorithmic complexity for short strings: A glance into the innermost structure of randomness. \emph{Appl Math Comput} 219(1):63--77.
\bibitem[Dessalles 2013]{dessalles} Dessalles, J-L. (2013), Algorithmic simplicity and relevance. In D. L. Dowe (Ed.), \emph{Algorithmic Probability and Friends. Bayesian Prediction and Artificial Intelligence: Papers from the Ray Solomonoff 85th Memorial Conference}, LNAI series.
\bibitem[Dodig-Crnkovic 2007]{gordana} Dodig-Crnkovic G. (2007), Where do New Ideas Come From? How do they Emerge? Epistemology as Computation (Information Processing). In C. Calude ed. \emph{Randomness \& Complexity, from Leibniz to Chaitin}.
\bibitem[Dowe and H\'ajek 1997]{dowe1} Dowe DL, H\'ajek AR (1997) A computational extension to the Turing test. \textit{Tech. Rep. 97/322}, nil, department of Computer Science, Monash University.
\bibitem[Dowe and H\'ajek 1998]{dowe2} Dowe DL, H\'ajek AR (1998) A non-behavioural, computational extension to the Turing Test. In \textit{Proc. Int. Conf. on Computational Intelligence and Multimedia Applications}, pp 101---
106, gippsland, Australia.
\bibitem[Gauvrit \& Morsanyi 2014]{gauvrit2014} Gauvrit N, Morsanyi K (2014) The Equiprobability Bias from a Mathematical and Psychological Perspective. \emph{Adv Cog Psy} 10(4):119-130
\bibitem[Gauvrit et al. 2014a]{gauvrit2014a} Gauvrit N, Zenil H, Delahaye J-P, et al (2014a) Algorithmic complexity for short binary strings applied to Psychology: a primer. \emph{Behav Res Methods} 46(3):732-744.
\bibitem[Gauvrit et al. 2014b]{gauvrit2014b} Gauvrit N, Singmann H, Soler-Toscano F and Zenil H. (2014b). Algorithmic complexity for psychology: A user-friendly implementation of the Coding Theorem Method. \textit{Behavior Research Methods}, DOI: 10.3758/s13428-015-0574-3 (online ahead of print).
\bibitem[Gauvrit et al. 2014c]{gauvrit2014c} Gauvrit N, Soler-Toscano F, Zenil H (2014c) Natural scene statistics mediate the perception of image complexity. \emph{Vis Cogn} 22(8):1084-1091.
\bibitem[G{\"o}del 1931]{godel} G{\"o}del K, (1931) {\"U}ber formal unentscheidbare S{\"a}tze der Principia Mathematica und verwandter Systeme, I. and On formally undecidable propositions of Principia Mathematica and related systems I in Solomon Feferman, ed., 1986. Kurt G{\"o}del Collected works, Vol. I. Oxford University Press: 144-195.
\bibitem[Hofstadter 2007]{hofstadter} Hofstadter, Douglas (2008). \emph{I Am A Strange Loop}. Basic Books.
\bibitem[Hsu 2010]{hsu} Hsu AS, Griffiths TL, Schreiber E (2010) Subjective randomness and natural scene statistics. \emph{Psychon B Rev} 17(5):624-629
\bibitem[Kahneman et al. 1982]{kahneman} Kahneman D, Slovic P, Tversky A (1982). \emph{Judgment under uncertainty: Heuristics and biases}. Cambridge University Press, Cambridge.
\bibitem[Kempe et al. 2015]{kempe} Kempe V, Gauvrit N, Forsyth D (2015) Structure emerges faster during cultural transmission in children than in adults. \emph{Cognition}, 136:247-254.
\bibitem[Kersten \& Earles 2001]{kersten} Kersten AW, Earles JL (2001) Less really is more for adults learning a miniature artificial language. \emph{J Mem Lang} 44(2):250-273.
\bibitem[Kirby et al. 2008]{kirby} Kirby S, Cornish H, Smith K (2008) Cumulative cultural evolution in the laboratory: An experimental approach to the origins of structure in human language. \emph{P Natl Acad Sci}, USA 105(31):10681-10686.
\bibitem[Kirchherr et al. 1997]{miracle} Kirchherr W, Li M, Vit\'anyi P (1997) The Miraculous Universal Distribution. \textit{The Mathematical Intelligencer}, vol. 19(4), pp. 7--15.
\bibitem[Kirk 1995]{kirk} Kirk R (1995) How is consciousness possible? In: Metzinger T (ed) \textit{Conscious Experience}, Ferdinand Schoningh (English edition published by Imprint Academic), pp 391--408
\bibitem[Klingberg et al. 2009]{jesper} Edin, F, Klingberg, T, Johansson, P, McNab, F, Tegn\'er, J, Compte, A. (2009) Mechanism for top-down control of working memory capacity, \emph{P Natl Acad Sci}, USA 106(16), 6802--6807.
\bibitem[Kolmogorov 1965]{kolmogorov} Kolmogorov AN (1965). Three Approaches to the Quantitative Definition of Information. \textit{Problems Inform. Transmission} 1 (1): 1--7.
\bibitem[Kryazhimskiy et al. 2014]{kryazhimskiy} Kryazhimskiy S, Rice DP, Jerison ER, Desai MM. (2014), Microbial evolution. Global epistasis makes adaptation predictable despite sequence-level stochasticity, \emph{Science}, Jun 27;344(6191):1519--22.
\bibitem[Lecoutre 1992]{lecoutre} Lecoutre MP (1992) Cognitive models and problem spaces in ``purely random'' situations. \emph{Educ Stud Math} 23(6):557-568.
\bibitem[Levin 1974]{levin} Levin LA (1974) Laws of information conservation (non-growth) and aspects of the foundation of probability theory. \textit{Problems Information Transmission}, 10(3):206-210.
\bibitem[Ma 2013]{ma} Ma L, Xu F (2013) Preverbal infants infer intentional agents from the perception of regularity. \emph{Dev Psychol} 49(7):1330.
\bibitem[Maguire et al. 2014]{maguire} Maguire, P., Moser, P., Maguire, R. \& Griffith, V. (2014) Is consciousness computable? Quantifying integrated information using algorithmic information theory. In P. Bello, M. Guarini, M. McShane, \& B. Scassellati (Eds.). \emph{Proceedings of the 36th Annual Conference of the Cognitive Science Society}. Austin, TX: Cognitive Science Society.
\bibitem[Mandler and Shebo 1982]{mandler} Mandler G, Shebo B J (1982) Subitizing: an analysis of its component processes. \emph{J Exp Psychol Gen} 111(1):1.
\bibitem[Mathy \& Feldman 2012]{mathy} Mathy F, Feldman J (2012) What's magic about magic numbers? Chunking and data compression in short-term memory. \emph{Cognition} 122(3):346-362
\bibitem[Mathy et al. 2014]{Mathy2014} Mathy F, Chekaf M, Gauvrit N (2014) Chunking on the fly in working memory and its relationship to intelligence. In: Abstracts of the 55th Annual meeting of the Psychonomic Society. Abstract \#148 (p. 32), University of California, Long Beach, 20--23 November 2014.
\bibitem[Matthews 2013]{matthews} Matthews W J (2013) Relatively random: Context effects on perceived randomness and predicted outcomes. \emph{J Exp Psychol Learn} 39(5):1642.
\bibitem[McDermott 2014]{drew} McDermott D (2014) On the Claim that a Table-Lookup Program Could Pass the Turing Test. \textit{Minds and Machines} 24(2), pp 143--188.
\bibitem[Miller 1956]{miller} Miller G. A. (1956) The magical number seven, plus or minus two: some limits on our capacity for processing information. \emph{Psychol Rev} 63(2):81
\bibitem[Oberauer 2004]{oberauer} Oberauer K, Lange E, Engle RW (2004) Working memory capacity and resistance to interference. \emph{J Mem Lang} 51(1):80-96.
\bibitem[Parberry 1997]{parberry} Parberry I (1997) Knowledge, Understanding, and Computational Complexity. In \textit{Optimality in Biological and Artificial Networks?},
Levine DS, Elsberry WR (eds.), chapter 8, pp. 125--144, Lawrence Erlbaum Associates.
\bibitem[Pearl 2012]{pearl} \url{http://blogs.wsj.com/digits/2012/03/15/work-on-causality-causes-judea-pearl-to-win-prize/} as accessed on Dec. 27, 2014.
\bibitem[Pepperberg 2006]{pepperberg} Pepperberg I M (2006) Grey parrot numerical competence: a review. \emph{Anim Cogn} 9(4):377-391.
\bibitem[Peng et al. 2014]{peng} Peng Z, Genewein T, and Braun DA (2014) Assessing randomness and complexity in human motion trajectories through analysis of symbolic sequences, \emph{Front Hum Neurosci.} 8: 168.
\bibitem[Penrose 1990]{penrose} Penrose R (1990). \textit{The emperor's new mind: Concerning computers, minds and the laws of physics}. London: Vintage.
\bibitem[Perlis 1990]{perlis} Perlis D (2005) Hawkins on intelligence: fascination and frustration. \textit{Artificial Intelligence}. 169:184--191.
\bibitem[Reznikova \& Ryabko 2011]{reznikova} Reznikova Z and Ryabko B (2011), Numerical competence in animals, with an insight from ants. \emph{Behaviour} 148, 405-434. 
\bibitem[Reznikova \& Ryabko 2012]{reznikova2} Reznikova Z and Ryabko B (2012). Ants and Bits. \emph{IEEE Information Theory Society} Newsletter March. 
\bibitem[Ryabko \& Reznikova 2009]{reznikova3} Ryabko B and  Reznikova Z (2009). The Use of Ideas of Information Theory for Studying ``Language'' and Intelligence in Ants. \emph{Entropy} 11, 836-853; doi:10.3390/e1104083.
\bibitem[Searle 1980]{searle} Searle J (1980) Minds, Brains and Programs, \emph{Behavioral and Brain Sciences,} 3: 417--57.
\bibitem[Smith \& Wonnacott 2010]{smith} Smith K, Wonnacott E (2010) Eliminating unpredictable variation through iterated learning. \emph{Cognition} 116(3):444-449.
\bibitem[Solomonoff 1964]{solomonoff} Solomonoff RJ (1964) A formal theory of inductive inference: Parts 1 and 2. \textit{Information and Control}, 7:1--22 and 224--254.
\bibitem[Soler-Toscano et al. 2014]{solerplos} Soler-Toscano F, Zenil H, Delahaye J-P and Gauvrit N (2014), Calculating Kolmogorov Complexity from the Output Frequency Distributions of Small Turing Machines, \emph{PLoS ONE} 9(5): e96223.
\bibitem[Spelke 2007]{spelke} Spelke ES, Kinzler KD (2007) Core knowledge. \emph{Dev Science,} 10(1):89-96.
\bibitem[Oizumi et al. 2014]{tononi} Oizumi M, Albantakis L, Tononi G (2014), From the Phenomenology to the Mechanisms of Consciousness: Integrated Information Theory 3.0. \emph{PLoS Computational Biology} 10 (5).
\bibitem[Turing 1938]{turing} Turing AM (1938), ``On Computable Numbers, with an Application to the Entscheidungsproblem: A correction'', \emph{Proceedings of the London Mathematical Society,} 2, 43 (6): 544--6, 1937.
\bibitem[Turing 1950]{turing2} Turing AM (1950), Computing Machinery and Intelligence, \emph{Mind LIX} (236): 433--460.
\bibitem[Xu \& Garcia 2008]{xu2008} Xu F, Garcia V (2008) Intuitive statistics by 8-month-old infants. \emph{P Natl Acad Sci.} USA 105(13):5012-5015.
\bibitem[Xu et al 2005]{xu} Xu F, Spelke ES, Goddard S (2005) Number sense in human infants. \emph{Dev Science}, 8(1):88-101.
\bibitem[T\'egl\'as et al. 2011]{teglas} T\'egl\'as E, Vul E, Girotto V et al (2011) Pure reasoning in 12-month-old infants as probabilistic inference. \emph{Science} 332(6033):1054-1059
\bibitem[Wang et al. 2014]{wang} Wang Z, Li Y, Childress AR, Detre JA (2014) Brain Entropy Mapping Using fMRI. \emph{PLoS ONE} 9(3): e89948.
\bibitem[Zenil et al. 2015]{kolmo2d} Zenil H, Soler-Toscano F, Delahaye J.-P., Gauvrit N. (2015). Two-dimensional Kolmogorov complexity and validation of the Coding Theorem Method by compressibility. \emph{PeerJ Comput. Sci.}, 1:e23.
\bibitem[Zenil et al. 2014]{zenilgraph} Zenil H, Soler-Toscano F, Dingle K, Louis AA (2014). Correlation of automorphism group size and topological properties with program-size complexity evaluations of graphs and complex networks. \emph{Physica A: Statistical Mechanics And Its Applications,} 404:341--358.
\bibitem[Zenil \& Hernandez-Quiroz 2007]{zenilmind} Zenil H and Hernandez-Quiroz F (2007), On the Possible Computational Power of the Human Mind. In \emph{Worldviews, Science and Us, Philosophy and Complexity,} Gershenson C, Aerts D, and Edmonds B. (eds), World Scientific.
\bibitem[Zenil 2013]{zenilbehavior} Zenil H (2013), Algorithmic Complexity of Animal Behaviour: From Communication to Cognition. \emph{Theory and Practice of Natural Computing Second International Conference Proceedings}, TPNC 2013. C\'aceres, Spain, December 3-5.
\bibitem[Zenil et al. 2012a]{entropy} Zenil H, Gershenson C, Marshall JAR and Rosenblueth D (2012), Life as Thermodynamic Evidence of Algorithmic Structure in Natural Environments, \emph{Entropy}, 14(11), 2173--2191. 
\bibitem[Zenil \& Delahaye 2010]{zenilalgo} Zenil H and Delahaye J-P (2010), On the Algorithmic Nature of the World. In Dodig-Crnkovic G and Burgin M (eds), \emph{Information and Computation,} World Scientific Publishing Company. 
\bibitem[Zenil \& Marshall 2013]{zenilubiquity} Zenil H and Marshall JAR (2013), Some Aspects of Computation Essential to Evolution and Life, \emph{Ubiquity} (ACM), vol. 2013, pp 1-16.
\bibitem[Zenil to appear]{zenilrobot} Zenil H (to appear), Quantifying Natural and Artificial Intelligence in Robots and Natural Systems with an Algorithmic Behavioural Test. In Bonsignorio FP, del Pobil AP, Messina E, Hallam J (eds.), Metrics of sensory motor integration in robots and animals, Springer.
\bibitem[Zenil 2014a]{zeniljetai} Zenil H (2014) Algorithmicity and Programmability in Natural Computing with the Game of Life as an In Silico Case Study, \emph{J Exp Theor Artif Intell.,} (online 02 Sep).
\bibitem[Zenil 2014b]{zenilphilo} Zenil H (2014), What is Nature-like Computation? A Behavioural Approach and a Notion of Programmability, \emph{Philosophy \& Technology,} vol 27:3, pp 399--421.
\bibitem[Zenil et al. 2015]{zenilmarshalltegner} Zenil H, Marshall JAR, T\'egner J, Approximations of Algorithmic and Structural Complexity Validate Cognitive-behavioural Experimental Results (submitted, preprint available at \url{http://arxiv.org/abs/1509.06338}).
\bibitem[Zenil 2010]{zenilca} Zenil H (2010), Compression-based Investigation of the Dynamical Properties of Cellular Automata and Other Systems, \emph{Complex Systems,} vol. 19, No. 1, pp. 1-28.
\bibitem[Zenil and Villarreal-Zapata 2013]{zenilchaos} Zenil H and Villarreal-Zapata E (2013), Asymptotic Behaviour and Ratios of Complexity in Cellular Automata Rule Spaces, \emph{International Journal of Bifurcation and Chaos,} vol. 13, no. 9.

\end{thebibliography}
\end{document}